\documentclass[10pt,twocolumn,letterpaper]{article}

\usepackage[pagenumbers]{cvpr} % To force page numbers, e.g. for an arXiv version

\usepackage{url}
\usepackage{multirow}
\usepackage{makecell}
\usepackage{booktabs}
\usepackage{graphicx}
\usepackage{comment}
\usepackage{verbatim}

\definecolor{cvprblue}{rgb}{0.21,0.49,0.74}
\usepackage[pagebackref,breaklinks,colorlinks,allcolors=cvprblue]{hyperref}

\def\paperID{21310} % *** Enter the Paper ID here
\def\confName{CVPR}
\def\confYear{2026}

\title{LMGenDrive: Bridging Multimodal Understanding and Generative World Modeling for End-to-End Driving} 
\author{
Hao Shao$^{1}$ \quad
Letian Wang$^{2}$ \quad
Yang Zhou$^{2}$ \quad
Yuxuan Hu$^{1}$ \\
Zhuofan Zong$^{1}$ \quad
Steven L. Waslander$^{2}$ \quad
Wei Zhan$^{3}$ \quad
Hongsheng Li$^{1}$ 
\\[10pt]
$^{1}$CUHK MMLab \quad
$^{2}$University of Toronto \quad
$^{3}$UC Berkeley
}
\begin{document}

\maketitle

\begin{abstract}

 Recent years have witnessed remarkable progress in autonomous driving, yet generalization to long-tail and open-world scenarios remains the primary bottleneck for large-scale deployment. To address this, one line of research explores LLMs and VLMs for their vision-language understanding and reasoning capabilities, equipping AVs with the ability to interpret rare and safety-critical situations when generating driving actions. In parallel, another line investigates generative world models to capture the spatio-temporal evolution of driving scenes, enabling agents to imagine and evaluate possible futures before acting. Inspired by human intelligence, which seamlessly unites understanding and imagination as a hallmark of AGI, this work explores a unified model that brings these two capabilities together for autonomous driving.
We present LMGenDrive, the first framework that unifies LLM-based multimodal understanding with generative world models for end-to-end closed-loop autonomous driving. Given multi-view camera inputs and natural-language instructions, our model generates both realistic future driving videos and corresponding control signals. By coupling an LLM with generative video capabilities, LMGenDrive gains complementary benefits: future video prediction enhances spatio-temporal scene modeling, while the LLM provides strong semantic priors and instruction grounding learned from large-scale pretraining. A progressive three-stage training strategy—ranging from vision pretraining to multi-step long-horizon driving—is proposed to further improve stability and performance.
The resulting model can also operate in two complementary modes: low-latency online planning and autoregressive offline video generation.
Experiments show that LMGenDrive significantly outperforms state-of-the-art methods on challenging closed-loop driving benchmarks, improving instruction following, spatio-temporal understanding, and robustness to rare scenarios. 
Our work not only sets a new state-of-the-art in autonomous driving, but also demonstrates that unifying multimodal understanding and generation provides a promising direction for building more generalizable and robust embodied decision-making systems.

\end{abstract}

\section{Introduction}

Remarkable progress in autonomous driving has been witnessed in recent years with an increasing number of commercial autonomous vehicles (AVs) deployed on public roads. Amidst this momentum, end-to-end autonomous driving has emerged as a particularly vibrant research direction. Unlike traditional modular pipelines that separately handle perception, prediction, and planning with handcrafted interfaces, end-to-end models provide a holistic paradigm with the potential to remove information bottlenecks among modules, better align model optimization with system-level performance, and scale effectively with large amounts of driving data.

Despite this progress, the problem of generalization remains the central bottleneck for the entire autonomous driving community. As we approach the frontier of real-world deployment, the ability to robustly handle long-tail edge cases and operate in open-world settings remains the defining challenge for AV systems. These scenarios can include rare but safety-critical events, distribution shifts across regions, adversarial weather conditions, as well as complex social interactions and ambiguous intent among agents. This challenge manifests across the autonomy stack: perception systems struggle to identify open-set entities, while prediction and planning models falter in extrapolating to nondeterministic and previously unseen behaviors. These generalization failures represent the last barrier between research prototypes and truly scalable, globally deployable autonomous vehicles.

Amid this backdrop, large language models (LLMs) have demonstrated strong capabilities in language understanding and structured decision modeling across a wide range of tasks. Recent models such as GPT-5 and DeepSeek-R1~\citep{guo2025deepseek} showcase robust capabilities in commonsense reasoning, abstraction, and decision-making. Meanwhile, vision-language models (VLMs) further extend this capacity to the multimodal domain, enabling unified interpretation of textual and visual inputs~\citep{wang2025internvl3,bai2025qwen2,liu2023visual,li2022blip,alayrac2022flamingo}. 
% with broad applications spanning from multi-hop question answering to real-world embodied tasks.
Inspired by these vision-language understanding and reasoning capabilities, a wave of research has begun exploring how to equip AV systems with LLMs and VLMs to address the open-world, long-tail challenges in autonomous driving. As exemplified in works such as LMDrive~\citep{shao2024lmdrive} and GPT-Driver~\citep{mao2023gpt}, these models act as the cognitive brains to interpret ambiguous scenarios and guide complex decision-making under uncertainty. However, most existing LLM- or VLM-empowered driving methods follow the paradigm that maps inputs directly to actions, falling short in explaining and capturing the temporal evolution of driving scenes—an essential factor for robust and anticipatory planning.
%Moreover, recent studies have shown that LLMs and VLMs often fall short in spatio-temporal understanding~\citep{chen2024spatialvlm, cheng2024videollama}.

Meanwhile, another stream of research, world modeling~\citep{ha2018world}, has emerged to simulate the spatio-temporal evolution of the scenes, as exemplified by video-based works such as Genie-3~\citep{ball2025genie} and Pandora~\citep{xiang2024pandora}. 
Their potential has also been actively explored in the autonomous driving domain, enabling the agent to ``imagine" different futures before committing to a plan.
However, existing works either focus on solely generating high-fidelity scenes~\citep{hu2023gaia,russell2025gaia,gao2023magicdrive,gao2024vista,ji2025cogen,wang2024drivedreamer,yang2024generalized,zhou2026drivinggen,shao2024visual}, or utilize world models as a plug-and-play forecasting module to rank multiple possible plans~\citep{wang2024driving,wang2025prophetdwm}. The integration of joint video generation and motion planning remains underexplored, limiting their ability to address critical challenges such as cumulative control errors, human-robot interaction, and the temporal consistency between generated actions and videos—factors that are essential for long-horizon problem solving in real-world systems.
Moreover, these models generally lack the rich knowledge priors and instruction-following capabilities provided by LLMs.
% limiting their potential for language-guided planning.

% planning-con

% focused purely on generating 

% later efforts have extended toward joint video generation and motion planning, aiming to improve safety and robustness by providing a temporally grounded planning substrate. However, most of these methods are limited to open-loop settings~\citep{wang2025prophetdwm}. The absence of closed-loop evaluation prevents them from addressing critical challenges such as cumulative control errors, human-robot interaction, and the temporal consistency of generated actions—factors essential for long-horizon problem sovling in real-world systems. 
% Moreover, these models generally lack the rich reasoning priors and instruction-following capabilities uniquely offered by LLMs, limiting their potential for language-guided planning.
% often using diffusion- or DiT-based frameworks. 

\begin{figure*}[t]
    \centering
    \includegraphics[width=1\linewidth]
    {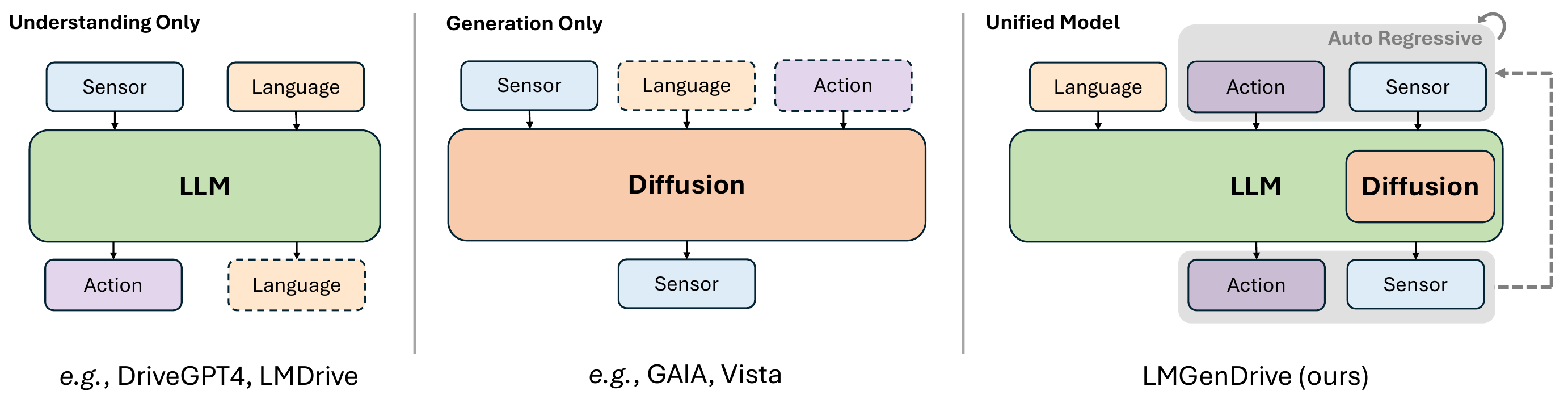}
    % \caption{Illustration comparing existing autonomous driving frameworks with ours. Our framework integrates LLM reasoning with built-in world models in an auto-regressive manner, enabling a more comprehensive understanding of the driving environment. In contrast, existing frameworks focus solely on either driving generation or understanding, lacking the ability to perform instruction-guided planning and spatio-temporal reasoning. Items marked with a dashed line are optional. The sensor suite includes RGB cameras, LiDAR, speed sensors, and more.}
    
    % Comparison between existing works and ours. 
    % Existing frameworks focus solely on vision-language scene understanding, or generative world modelling.
    % Our framework integrates LLM reasoning with built-in world models in an auto-regressive manner, enabling comprehensive understanding of the driving environment. 
    % either scene generation or understanding, lacking support for instruction-guided planning and spatio-temporal reasoning. 
    % Items marked with dashed lines are optional. The sensor suite includes RGB cameras, LiDAR, speed sensors, and others.
    \caption{
    Comparison between existing works and ours. Prior works either leverage LLMs/VLMs for multimodal understanding and reasoning, or adopt world models for video-based scene imagination, but treat these capabilities in isolation. In contrast, our proposed \textbf{LMGenDrive} unifies both within a closed-loop end-to-end framework: the LLM interprets and reasons over multimodal inputs, while the world model simulates future scene evolution, together enabling instruction-guided planning, spatio-temporal understanding, and robust long-horizon driving. 
    }
    % Items marked with dashed lines are optional, and the sensor suite includes RGB cameras, LiDAR, speed sensors, and others.
    \label{fig:motiv}
\end{figure*}

%  While 

% % In this context, our insight is that, 

% In this context, one research question naturally emerges, whether a unified multimodal understanding and generation could benefit each other, rushing on the path to AGI

% Thus, the research question pursued in this paper is that, whether a unified multimodal understanding and generation could benefit each other, rushing on the path to AGI. While existing works have demosntrated promising results in jointly optimizing generation and understandind benchmarks with a crafted unified architecture, whether it is the case for the domain of embodied agent and autonomous driving remain unclear.
In contrast to existing models that specialize in either understanding or generation, human cognition naturally integrates perception and imagination to support long-horizon decision-making. Inspired by this observation, we explore whether combining multimodal understanding with generative world modeling can improve robustness and temporal consistency in autonomous driving systems.
While recent studies~\citep{deng2025emerging,shi2024lmfusion,chen2025blip3,liao2025mogao} have shown encouraging results combining multimodal understanding and generation within a single model, whether this principle extends to embodied agents—and autonomous driving in particular—remains an open challenge.
In this work, we propose LMGenDrive, the first framework that unifies LLM-based multimodal understanding with generative world models for closed-loop end-to-end autonomous driving. 
Our unified model takes multi-view camera data and natural-language driving instructions as inputs, generating both multi-view future driving videos and control signals for the following timesteps. 
Concretely, we integrate an LLM and a diffusion-based video generation model: the LLM interprets and fuses visual observations with language instructions, producing learnable queries that capture the evolving scene states, which then serve as conditioning signals for the diffusion model to generate realistic multi-view driving futures.
Within this unified architecture, video generation enhances the LLM’s spatio-temporal understanding, while the LLM provides instruction-following and reasoning capabilities to the world model, together enabling more robust closed-loop driving.

% To establish reasoning and generative capabilities within LMGenDrive, we integrate a pretrained LLM and a diffusion-based video generation model. 
% For predictive planning, we introduce learnable queries along with learnable input/output adapters. 
% For driving video generation, we use learnable queries to obtain a sequence of intermediate visual features representing changes in the driving environment. 
% To reconstruct a coherent 3D world scenario from the multi-view camera inputs, we propose a dedicated world generator. 
% This model combines the generated world-change features with the current world state—represented by the latest multi-view camera data—to predict high-level multi-view features for the next timestep. 
% These predicted features serve as conditioning inputs for the diffusion model, facilitating the generation of multi-view driving videos.

% We also propose a sequential three-stage training strategy.
% First, the vision encoder is pretrained with perception heads on large-scale driving data for robust scene understanding. Next, we integrate the frozen vision encoder with the LLM and video generator, and fine-tune them on a single-step prediction task to learn grounded instruction following and immediate action outcomes. Finally, we extend training to multi-step sequences, strengthening long-horizon reasoning and temporal modeling for continuous, long-term driving scenarios. This progressive approach enhances both overall performance and training stability.

To enable such a unified model, we also propose a three-stage curriculum training strategy for enhanced performance and stability. 
First, we pretrain a vision encoder for robust driving scene understanding. Next, the frozen encoder is integrated with the LLM and video generator, and fine-tuned on single-step prediction to ground instruction following and immediate action outcomes. Finally, training is extended to multi-step sequences, enhancing long-horizon temporal modeling and consistency and temporal modeling for continuous driving scenarios. 
% This progressive design improves both performance and stability.
Once trained, the model can be applied in two modes: (1) Online planning mode: the model solely predicts planning outputs, with the diffusion generation component discarded to reduce latency; (2) Offline data generation mode: the model conducts autoregressive video generation, where the generated video and predicted control signal serve as input for the next timestep, enabling extended and consistent driving video sequences.

To sum up, our contributions are threefold:
\begin{itemize}
    \item \textbf{Unified closed-loop framework.} 
    We present LMGenDrive, the first framework that unifies LLM-based multimodal understanding with generative world models for closed-loop end-to-end autonomous driving, bridging perception and imagination within a single architecture.
    
    \item \textbf{Progressive training and dual modes.} 
    We introduce a three-stage training pipeline, from vision pretraining to long-horizon multi-step driving, and support two usage modes: low-latency online planning and offline autoregressive video generation.
    
    \item \textbf{State-of-the-art performance and empirical insights.}
    LMGenDrive achieves state-of-the-art closed-loop performance on challenging autonomous driving benchmarks, improving instruction following, spatio-temporal reasoning, and robustness to long-tail scenarios. These results highlight the complementary benefits of combining multimodal understanding with generative world modeling.
\end{itemize}

\section{Related works}

\subsection{End-to-end driving}
Much progress has been made in end-to-end autonomous driving, with many recent methods based on imitation learning. UniAD~\citep{hu2023planning} unified full-stack driving tasks through query-based interfaces, while ThinkTwice~\citep{jia2023think} retrieved critical-region information to refine predictions~\citep{zhou2024smartrefine}. InterFuser~\citep{shao2023safety} used transformers to fuse multi-modal, multi-view sensor data for richer scene understanding. ReasonNet~\citep{shao2023reasonnet} leveraged both temporal and global information of the driving scene to enhance perception, particularly in occlusion scenarios. Para-Drive~\citep{weng2024drive} proposed a fully parallel architecture with a shared BEV representation, and DriveTransformer~\citep{jia2025drivetransformer} went further by discarding BEV features and using pure transformers to aggregate sensor and query information. 
Diffusion models have also been explored for modeling diverse driving behaviors.
DiffusionPlanner~\citep{zheng2025diffusion} learns a diffusion-based driving policy,
while DiffAD~\citep{wang2025diffad} formulates perception and decision-making as
conditional image generation. More recent frameworks, such as SimLingo~\citep{renz2025simlingo}
and Orion~\citep{fu2025orion}, incorporate large language models or generative world
modeling into the driving loop, enabling richer semantic grounding and action-conditioned
scene reasoning. Despite these advances, most approaches still struggle with rare corner
cases and lack the reasoning ability needed to generalize beyond the training distribution.

%

% 然而，这些方法往往在long-tail corner cases 中struggle，没有reasoninging能力，难以处理training data以外的情况。

% refoa novel diffusion probabilistic model that redefines autonomous driving as a conditional image generation task. By rasterizing heterogeneous
% targets onto a unified bird’s-eye view (BEV) and modeling their latent distribution, DiffAD unifies various driving objectives and jointly optimizes all driving tasks in a single framework, significantly reduces system complexity
% and harmonizes task coordination. The reverse process iteratively refine the generated BEV image, resulting in more robust and realistic driving behaviors.

% fully harness the power of diffusion models with a specifically designed architecture for high-performance motion planning, without overly reliant on rule-based refinement.
% can achieve personalized driving behavior at runtime by utilizing a flexible guidance mechanism, 

% However, these end-to-end methods lack the ability to verbally or textually interact with humans (passengers), and usually have low explanatility in the decision-making process.

% Referring to LMdrive

% Others:
% PARA-Drive: Parallelized Architecture for Real-time Autonomous Driving

% DriveTransformer: Unified Transformer for Scalable End-to-End Autonomous Driving

% DiffAD: A Unified Diffusion Modeling Approach for Autonomous Driving

% DiffE2E: Rethinking End-to-End Driving with a Hybrid Action Diffusion and Supervised Policy

\subsection{MLLM for autonomous driving}
Recent advances in large language models (LLMs)~\citep{guo2025deepseek,yang2025qwen3,touvron2023llama1,touvron2023llama2,jaeger2023hidden,jia2023driveadapter} and vision–language models (VLMs)~\citep{bai2025qwen2,zhu2023minigpt,liu2023visual,wang2025internvl3,shao2024visual,liu2024sphinx,qian2021blending,zong2024mova} have motivated integrating MLLMs into autonomous driving for stronger reasoning and explainability. Early works like GPT-Driver~\citep{mao2023gpt} and LanguageMPC~\citep{sha2023languagempc} convert driving scenes into textual inputs for direct reasoning. LLM-Driver~\citep{chen2023driving} utilized the numeric vector as the input modality and fused vectorized object-level 2D scene representation to enable the LLM to answer the questions based on the current environment. Later methods employ VLMs to process images and videos: some focus on visual question answering for scene understanding and optional action output (\textit{e.g.}, DriveLM~\citep{sima2024drivelm}, DriveGPT4~\citep{xu2023drivegpt4}, DriveVLM~\citep{tian2024drivevlm}), while others predict driving actions end-to-end (\textit{e.g.}, LMDrive~\citep{shao2024lmdrive}, DriveMoE~\citep{yang2025drivemoe}, BEVDriver~\citep{winter2025bevdriver}). Agentic designs with hierarchical control, tool use, and memory, such as Agent-Driver~\citep{mao2023language} and AD-H~\citep{zhang2024ad}, further extend capabilities.
However, most MLLM-based approaches emphasize planning or explanation and lack robust modeling of how scenes and surrounding objects evolve over time—a key requirement for anticipating events and ensuring safe, long-horizon decision-making.

\subsection{World models for autonomous driving}
The concept of a \textit{world model}, a predictive model that simulates environment dynamics, has regained attention. Video generation has become a leading paradigm, supported by advances in generative modeling, large-scale video datasets, and with wide applicability. In autonomous driving, temporally grounded video prediction provides rich context for understanding and decision-making.
Several methods treat pure video generation as world modeling. GAIA~\citep{hu2023gaia} conditions generation on image, text, and action inputs. GAIA-2~\citep{russell2025gaia} extends this to multi-view scenes, and MagicDrive~\citep{gao2023magicdrive} adds control signals such as HD maps and bounding boxes. Vista~\citep{gao2024vista} scales to internet-scale driving data, while CoGen3D~\citep{ji2025cogen} predicts 3D-consistent representations before video synthesis to improve spatial coherence.
Beyond pure generation, DriveWM~\citep{wang2024driving} predicts alternative futures for conditional planning. More recent work—DriveDreamer~\citep{wang2024drivedreamer}, GenAD~\citep{yang2024generalized}, and ProphetDWM~\citep{wang2025prophetdwm}—jointly models videos and actions but still mainly uses open-loop settings without direct feedback.
The most related work is LAW~\citep{li2024enhancing}, which combines world modeling with closed-loop planning. However, it supervises the world model only with next-frame hidden features rather than full video generation, limiting its ability to simulate realistic scenes or generate synthetic data. Moreover, it lacks language model integration and therefore cannot support instruction following or natural language grounding in driving. To our knowledge, this is the first framework to unify LLM-based reasoning with video world models for closed-loop end-to-end autonomous driving.

\begin{figure*}
    \centering
    \includegraphics[width=1.0\linewidth]
    {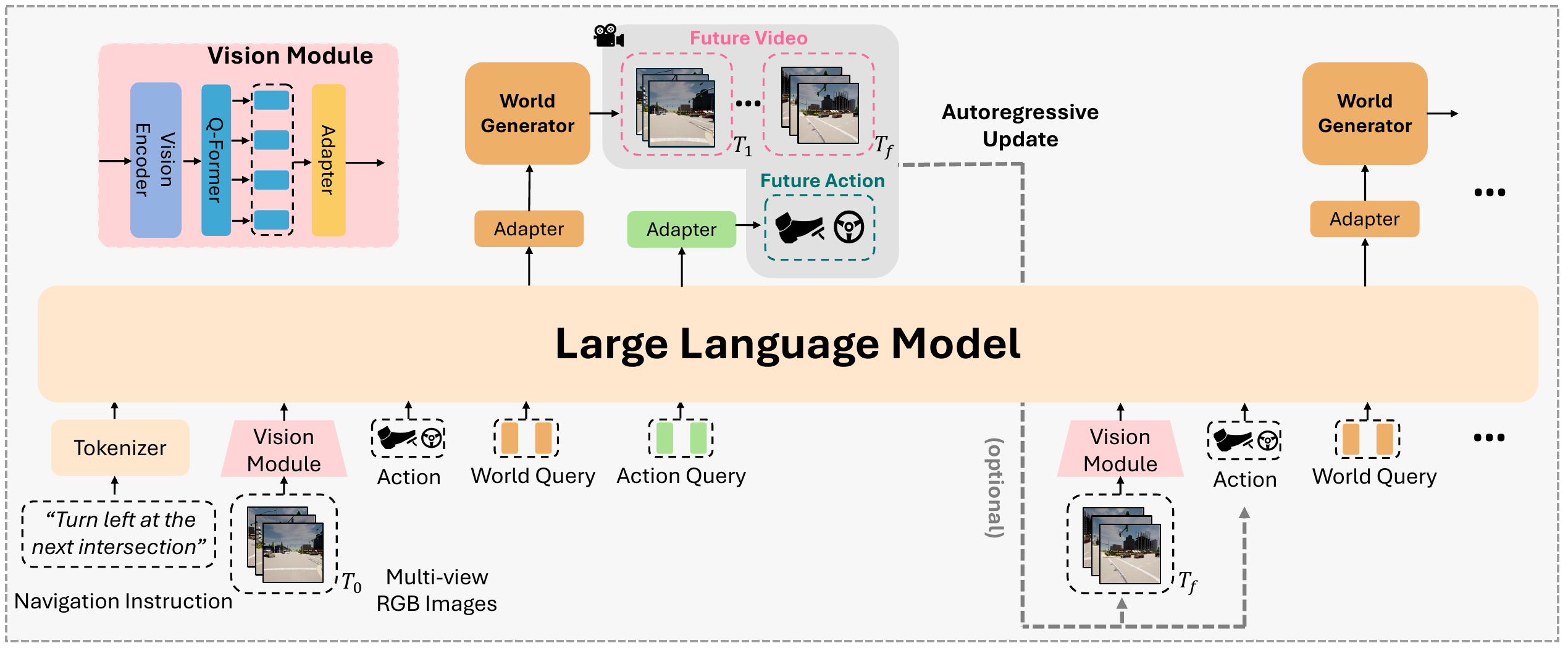}
    \caption{Overview of our unified understanding and generation architecture. We start by encoding the language instruction, multi-view RGB images, and the current action into the LLM. Two sets of learnable queries, world query and action query, are then fed into the LLM, and ultimately used to generate the future driving video and corresponding actions. The framework supports two operation modes: (1) offline data generation mode, an autoregressive generation process is adopted, where the last frame of the future video and the predicted action are used as inputs for the next timestep; (2) online planning mode, real-world data are provided as inputs for the following timestep.
    }
    \label{fig:framework}
\end{figure*}

\section{Method}
% \textit{Generation part, refer to DreamLLM, Pandora: Towards General World Model with Natural Language Actions and Video States , minigpt-5}

% \textit{Others: Refer to LMdrive}

\subsection{Overall Framework}

In this work, we propose LMGenDrive, a framework that unifies textual understanding/reasoning, future scene generation, and end-to-end planning.
As illustrated in Figure~\ref{fig:framework}, LMGenDrive is composed of three major components: (1) a vision encoder that processes multi-view camera sensor data for scene understanding and generating visual tokens; (2) a large language model and its associated component (tokenizer, Q-Former, and adapters) that takes in the language instruction, input visual tokens, world queries, and action queries, to predict the future driving scenes and actions;
(3) a multi-view world generator that takes future scene tokens from the LLM and multi-view images from the last frame as inputs, to generate future multi-view driving videos.
We will introduce the vision encoder in Section~\ref{sec:visual_encoder}, the LLM with its associated components in Section~\ref{sec:llm}, and the multiview world generator in Section~\ref{sec:video_generator}.
Finally, we describe the training recipe in Section~\ref{sec:training_setup}.

\subsection{Vision encoder}
\label{sec:visual_encoder}

% Inspired by previous works ~\citep{shao2024lmdrive, jaeger2023hidden}, we design a vision encoder to process and fuse multi-view camera sensor data. 
The vision encoder is designed to perceive the environment by processing, fusing, and transforming sensor data into visual tokens that can be consumed by the language model.
Prior works~\citep{shao2024lmdrive, jaeger2023hidden} typically leverage both multi-view images and LiDAR sensor inputs, where the LiDAR inputs are encoded into bird’s-eye view (BEV) queries to extract information from multi-view images. However, our setting focuses on autoregressive video generation—where LiDAR is only available at the current frame but not in future frames. As a result, we replace LiDAR inputs with BEV positional encodings, enabling effective perception while maintaining compatibility with future video generation.
The vision encoder consists of three parts:
(1) In the sensor encoding part, for each image input, a 2D backbone Resnet~\citep{he2016deep} is applied to extract the image feature map, which is flattened to one-dimensional tokens. Tokens from different views are then fused by a transformer encoder.
(2) In the BEV decoder, BEV position encodings serve as $H \times W$ queries to attend to the multi-view image features and generate BEV tokens. In addition, the learnable queries and one extra query generate corresponding waypoint tokens and one traffic-light token, respectively.
The three types of visual tokens (BEV, waypoint, and traffic light) will be presented to the LLM, providing rich scene information.
(3) Lastly, as the first-stage training, the vision encoder is pretrained on perception tasks (BEV object detection, traffic light recognition, waypoint prediction) by feeding the three types of tokens to additional prediction heads. Three loss terms, including the detection loss~\citep{shao2023reasonnet}, the $l_1$ waypoint loss and the cross-entropy traffic light prediction loss, are applied respectively. Note that, following LMDrive~\citep{shao2024lmdrive}, once pretrained, these prediction heads are discarded and the encoder is frozen, serving as the vision encoder for the large language model.

\subsection{LLM for instruction-following driving and scene understanding}
\label{sec:llm}

As depicted in Figure \ref{fig:framework}, our system casts the LLM as the ``brain” of the entire driving pipeline: it ingests sensor tokens emitted by the frozen vision encoder at every frame and parses natural-language commands, to forecast upcoming maneuvers and emits conditioning features for subsequent video generation.
We adopt LLaMA~\citep{touvron2023llama1} as the linguistic architecture due to its broad success in both language-centric~\citep{zheng2023judging,geng2023koala} and vision-grounded~\citep{liu2023visual,zhu2023minigpt} instruction-tuning settings.
% To interface the LLM with instructions, perceptual inputs, and action outputs, the architecture incorporates three auxiliary modules: (1) a tokenizer, (2) a Q-Former, and (3) a pair of lightweight adapters.

% \vspace{1mm}
\noindent \textbf{Instruction and visual tokenization.} As the model takes navigation instruction and multi-view image as inputs, their tokenization is our first step. 
For the navigation instruction, we tokenize them with the LLaMA tokenizer~\citep{touvron2023llama1}. For the multi-view images, each frame is tokenized by the aforementioned vision encoder, and the resulting tokens are buffered together with the most recent token history (up to $T_{\text{max}}$ frames) to curb cumulative error and maintain temporal coherence during executing the driving instruction in the closed loop.
For each frame, the pretrained vision encoder outputs $H \times W$ BEV tokens, 4 waypoint tokens, and 1 traffic-light token. Passing all visual tokens (about 2k per frame) to the LLM is computationally prohibitive. To compress them, we use a Q-Former with 8 learnable queries per frame that attend to the raw tokens and distill them into compact frame-level features. An MLP adapter then projects these features to the LLM’s embedding dimension for seamless fusion with language tokens.

% \vspace{1mm}
\noindent \textbf{Action prediction.}
Together with the instruction and visual tokens, we feed $N$ learnable action query tokens into the LLM. After passing through the LLM and associated adapters, these queries evolve into $N$ latent feature vectors, each encoding the spatio-temporal context needed for motion planning. A subsequent two-layer MLP maps these $N$ feature vectors to $N$ predicted waypoints and outputs a binary flag indicating whether the current instruction has been completed. Finally, the predicted waypoints are converted into low-level control commands—brake, throttle, and steering—through two independent PID controllers~\citep{chen2020learning} that separately regulate longitudinal velocity and lateral heading, ensuring accurate trajectory tracking.

\begin{figure*}[!t]
    \centering
    \begin{minipage}[t]{0.65\textwidth}
    \includegraphics[width=\linewidth]{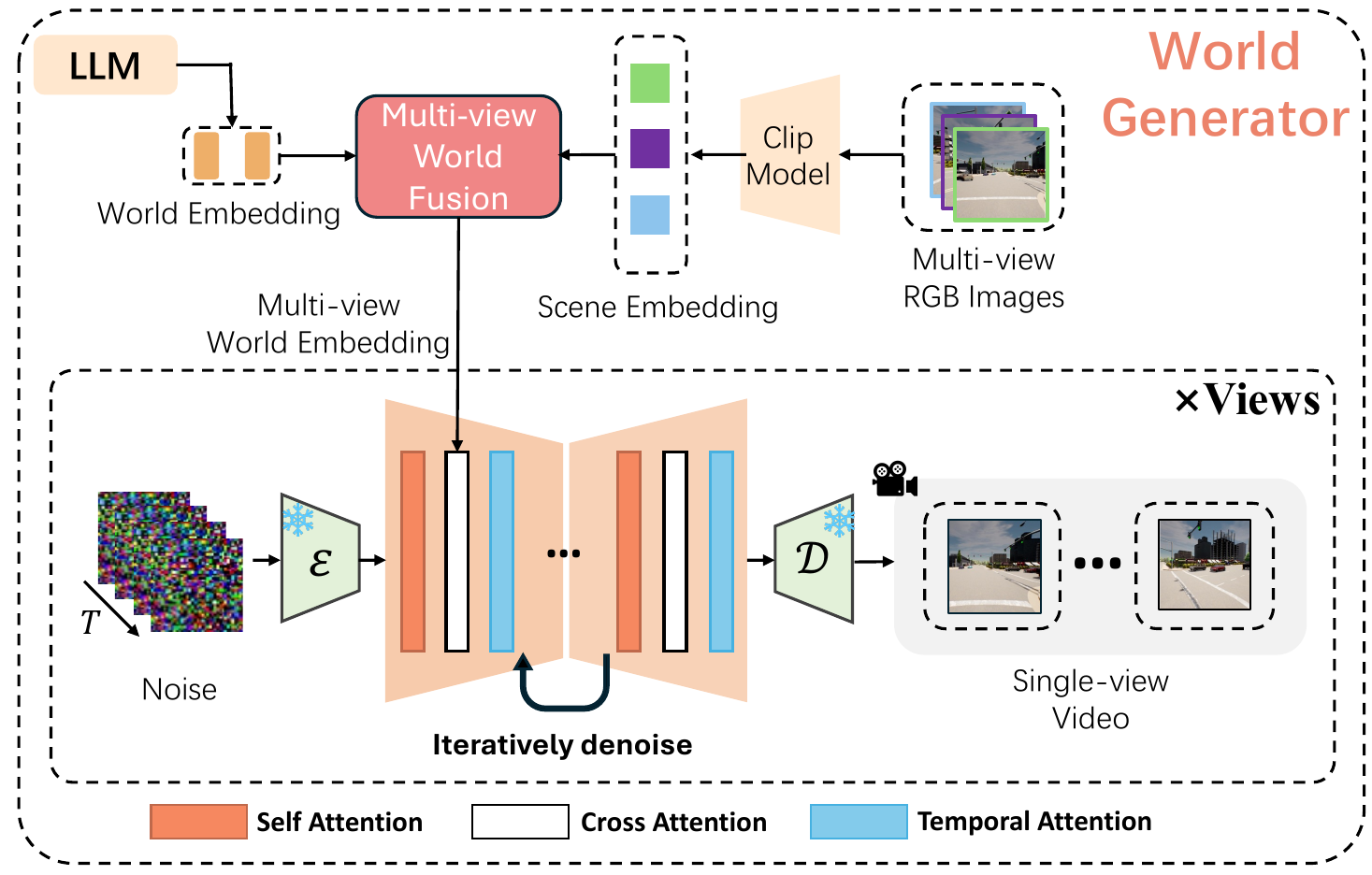}
    \end{minipage}

    \caption{Architecture of our world generator. We begin by fusing the world embedding obtained from the LLM with multi-view RGB images. The fused multi-view world embedding is then injected into the diffusion model with the cross-attention mechanism to generate multi-view future videos.}
    \label{fig:diffusion}
\end{figure*}

\subsection{Multi-View World Model}
\label{sec:video_generator}
While the LLM is responsible for instruction following and reasoning, autonomous driving also requires modeling the visual dynamics of the environment. To this end, we introduce a multi-view world model that generates future video frames conditioned on the LLM outputs. By aligning action predictions with video generation, our framework jointly reasons about both the agent’s behavior and the evolution of the surrounding world.
Our video generation process additionally \emph{supports} an autoregressive mode~\citep{xiang2024pandora}, which can be optionally enabled during inference: the future action and frame predicted at the previous timestep can be fed back into the LLM as input for the next prediction.

\noindent \textbf{World Query Conditioning.} 
As shown in Figure~\ref{fig:framework}, in addition to the instructional tokens, visual tokens, and action query mentioned above, the LLM also takes the world query as input. Passing through the LLM, these world queries aggregate information from instructions, sensor inputs, and the actions, thereby enabling the model to form an internal representation of the world dynamics. 
Conceptually, these queries serve as a bridge to the world’s temporal evolution, and act as the conditioning signal fed into the following world generator to synthesize future driving videos.

% Then the queried features are fed into the world generator to generate future driving videos. 

% Note that at the first time step, the input considers a single-frame multi-view images, while at the later time step, the inputs consist of a historic window of multi-view images. 

\noindent \textbf{Multi-view Image Conditioning.}
As shown in Figure~\ref{fig:diffusion}, besides using scene queries to capture world evolution, we incorporate the last-frame multi-view images to supply fine-grained appearance details and the initial world state. These images are encoded by a CLIP model into semantically rich features that emphasize visual textures and appearance. During end-to-end training, these CLIP features are fused with LLM representations through attention blocks. Self-attention aggregates and aligns multi-view information into a unified space, and cross-attention injects LLM guidance. This design not only provides an appearance prior for consistent video generation but also encourages the LLM to focus on dynamic, motion-related representations.

\noindent \textbf{World Generator.}  
After the multi-view world fusion step, we obtain a set of multi-view world embeddings, each corresponding to one camera view. Taking these embeddings as the final conditioning feature, our world generator employs a U-Net~\citep{ronneberger2015u} diffusion architecture to produce future frames. For each view, the associated embedding is injected into the diffusion process through a dedicated cross-attention module, ensuring that view-specific information is effectively transferred. The model follows the standard denoising diffusion process~\citep{ho2020denoising}: starting from pure Gaussian noise and progressively removing noise to generate a future video sequence. The output is a video tensor of shape $\mathbb{R}^{v \times t \times h \times w \times 3}$, where $v$ is the number of views, $t$ is the temporal length, and $h, w$ are the spatial resolution. Inspired by \citep{wang2024driving,guo2023animatediff}, we further augment the U-Net blocks with spatio-temporal transformers to better capture temporal dynamics and spatial structure in driving scenes.

\subsection{Training Recipe}
\label{sec:training_setup}

We adopt a three-stage training strategy to progressively build the model’s perception, reasoning, and generation capabilities. This curriculum ensures stable convergence and enables effective long-horizon temporal modeling.

\noindent \textbf{Stage 1: Vision Encoder Pretraining.}  
We first pretrain the vision encoder on single-frame perception tasks using 3M expert-collected frames generated in the CARLA simulator~\citep{shao2023safety}. Perception heads are attached for object detection, traffic light classification, and waypoint regression. After convergence, only the vision encoder is retained and frozen in later stages.

\noindent \textbf{Stage 2: Single-Step Planning and Generation.}  
Next, we jointly fine-tune the LLM and the video generator for single-step prediction. The vision encoder is frozen to reduce memory usage. The model takes as input a single-frame multi-view image, natural language instruction, action queries, and world queries, to predict the next waypoint, the instruction completion flag, and the future driving video. This stage enables the LLM to learn grounded instruction-following and understand how the world evolves under given actions. Simultaneously, the world generator learns to synthesize multi-view driving videos conditioned on the last frame and LLM-generated features.

\noindent \textbf{Stage 3: Multi-Step Long-Horizon Training.}  
We progressively expand training to 2–3-step sequences to strengthen long-horizon reasoning. Specifically, previously generated videos are autoregressively fed as input for the next step’s generation. To save memory, the video generator is frozen while gradients still propagate, and the LLM remains fully trainable. This design encourages the LLM to capture temporal dependencies—such as other agents’ intentions, speed, and interactions—over extended observation windows. As a result, the LLM develops stronger temporal abstraction and inductive reasoning abilities for dynamic driving scenes.

\noindent \textbf{Training Objectives.}  
We apply three loss terms in the last two stages:  
(1) $l_1$ waypoint regression loss;  
(2) binary classification loss for instruction completion;  
(3) diffusion loss for video generation:
\[
\mathcal{L}_{\text{DM}} = \mathbb{E}_{t, \epsilon} \left[\left\| \epsilon_\theta(\mathbf{z}_t, \mathbf{c}, t) - \epsilon \right\|^2\right],
\]
where $\mathbf{z}_t$ is the noisy latent at timestep $t$, $\epsilon$ is the added Gaussian noise, and $\mathbf{c}$ denotes conditioning features from the multi-view image and scene queries.

\section{Experiments}

\begin{table*}[h!]
\centering
\caption{Performance comparison on the LangAuto benchmark. We report the metrics for 3 evaluation runs. AD-H$\dagger$ leverages an extra model OPT-350M~\citep{zhang2022opt} for low-level control.}
\label{table:sota_comparasion}
\resizebox{1\textwidth}{!}{
\begin{tabular}{lccccccccc}
\toprule
\multirow{2}{*}{Methods} & \multicolumn{3}{c}{LangAuto} & \multicolumn{3}{c}{LangAuto-Short} & \multicolumn{3}{c}{LangAuto-Tiny} \\ \cmidrule(r){2-4} \cmidrule(r){5-7} \cmidrule(r){8-10}
                        &  DS $\uparrow$  & RC $\uparrow$   &  IS $\uparrow$      &  DS $\uparrow$  & RC $\uparrow$   &  IS $\uparrow$  &  DS $\uparrow$  & RC $\uparrow$   &  IS $\uparrow$   \\ \cmidrule(r){1-1} \cmidrule(r){2-4} \cmidrule(r){5-7} \cmidrule(r){8-10}
                  LMDrive~\citep{shao2024lmdrive}      & 10.7$\pm$3.8   &  16.2$\pm$4.9  &  0.63$\pm$0.04  &  14.2$\pm$4.4    &  20.1$\pm$4.4    &  0.72$\pm$0.04   &  20.1$\pm$4.1    &  24.7$\pm$5.1    &  0.75$\pm$0.03     \\
                  AD-H$\dagger$~\citep{zhang2024ad}     & 44.0 & 53.2 & 0.83 & 56.1 & 68.0 & 0.78 & 77.5 & 85.1 & 0.91   \\ 
                  
                  BEVDriver~\citep{winter2025bevdriver}     &  48.9 & 59.7 & 0.82 & 66.7 & 77.8 & 0.87 & 70.2 & 81.3 & 0.87  \\ 

                Ours & \textbf{62.2}$\pm$\textbf{3.3} & \textbf{74.5}$\pm$\textbf{4.1} & \textbf{0.85}$\pm$\textbf{0.04}
                & \textbf{77.1}$\pm$\textbf{4.1} & \textbf{87.9}$\pm$\textbf{3.5} & \textbf{0.88}$\pm$\textbf{0.03}
                & \textbf{84.1}$\pm$\textbf{3.6} & \textbf{92.5}$\pm$\textbf{4.0} & \textbf{0.92}$\pm$\textbf{0.04} \\

                        \bottomrule
\end{tabular}}

\end{table*}
% OPT350m]
\begin{table*}[ht]
\centering
\begin{minipage}[t]{0.58\linewidth}
\centering
\caption{Ablation study on the module design for planning performance.}
\label{tab:ablation-planning}
\resizebox{\linewidth}{!}{
\begin{tabular}{cccc}
\toprule
Module design  &  DS $\uparrow$  & RC $\uparrow$   &  IS $\uparrow$       \\ \cmidrule(r){1-1} \cmidrule(r){2-4} 
baseline      & \textbf{62.2}$\pm$\textbf{3.3} & \textbf{74.5}$\pm$\textbf{4.1} & \textbf{0.85}$\pm$\textbf{0.04} \\
w/o world generator    & 53.4$\pm$2.2  & 65.8$\pm$4.2  & 0.80$\pm$0.01    \\ 
w/o action queries     & 58.7$\pm$3.1  & 70.4$\pm$3.7  & 0.84$\pm$0.02 \\ 
w/o visual pre-training & 54.9$\pm$4.5  & 67.1$\pm$4.5  & 0.81$\pm$0.02 \\ 
w/o stage-3 training   & 55.6$\pm$4.5  & 68.9$\pm$4.5  & 0.80$\pm$0.02 \\
\bottomrule
\end{tabular}}
\end{minipage}%
\hfill
\begin{minipage}[t]{0.40\linewidth}
\centering
\caption{Ablation study on the module design for generation performance.}
\label{tab:ablation-generation}
\resizebox{\linewidth}{!}{
\begin{tabular}{ccc}
\toprule
Module design  &  FID$\downarrow$  &  FVD$\downarrow$       \\ \cmidrule(r){1-1} \cmidrule(r){2-3}
baseline      & \textbf{6.3}  & \textbf{286}  \\
w/o multi-view fusion    & 7.8   & 371      \\ 
world queries: 64 $\rightarrow$ 32     & 10.1   & 318  \\ 
world queries: 64 $\rightarrow$ 16     & 11.6   & 424  \\ 
\bottomrule
\end{tabular}}
\end{minipage}
\end{table*}

\subsection{Experiment Setup}
\textbf{Training Details.} 
During training, three synchronized RGB cameras (left, front, right) are resized to 224$^2$ pixels and sampled at 10\,Hz, and an 8-frame temporal window is considered.  
The network is tasked with predicting four future waypoints at $t+\{0.2,0.4,0.6,0.8\}\,\mathrm{s}$, along with eight future video frames from $t+0.1\,\mathrm{s}$ to $t+0.9\,\mathrm{s}$ in 0.1-second increments.  
We optimize the model using AdamW optimizer~\citep{loshchilov2018decoupled} with an initial learning rate of $1\times10^{-5}$ on eight NVIDIA H800 GPUs under DeepSpeed ZeRO-2; convergence is reached in roughly two days.  
Due to GPU-memory constraints, the third training stage operates on one to three timesteps. We allocate a total of 50k optimization iterations across the three stages, using 20k/20k/10k iterations for Stage-1/Stage-2/Stage-3,
respectively.
The system uses Vicuna-7B~\citep{chiang2023vicuna} as the LLM backbone, Stable Diffusion 1.5~\citep{rombach2022high} for image generation, and AnimateDiff~\citep{guo2023animatediff} for temporal modeling.

\noindent \textbf{Benchmark.}
We implement and evaluate our approach in the open-source CARLA simulator (version 0.9.10.1)~\citep{dosovitskiy2017carla} on the LangAuto benchmark~\citep{shao2024lmdrive}. The LangAuto benchmark comprises test routes spanning eight CARLA towns, diverse weather conditions, and deliberately misleading linguistic cues. It includes three tracks—LangAuto, LangAuto-Short, and LangAuto-Tiny—which differ primarily in route length.
During evaluation, each method acts as an agent that controls the vehicle using only natural-language commands and visual observations, and directly outputs the corresponding control signals to interact with the real-time environment. Consistent with LMDrive~\citep{shao2024lmdrive}, we report results on each track individually.

% We conduct standard closed-loop evaluations using CARLA simulator Dosovitskiy et al. [2017] on
% the LangAuto Benchmark Shao et al. [2023]. 

\noindent \textbf{Metric.}
Following the CARLA Leaderboard~\citep{leaderboard} and LangAuto~\citep{shao2024lmdrive}, we report route completion (RC), infraction score (IS), and driving score (DS). RC measures the fraction of the planned route completed before exceeding the deviation tolerance. IS penalizes collisions and traffic-rule violations via a decaying factor. DS, the product of RC and IS, serves as the primary overall indicator of safe and efficient driving.
For generated videos, we further evaluate perceptual quality using Fréchet Video Distance (FVD) and Fréchet Inception Distance (FID), which assess temporal consistency and visual realism, respectively.

% input sensor h,w=224, t=8, v=4(left, front, right)
% adamw, 1e-5
% 8 H800,zero2, for about 2 days training
% data sample rate: 10hz
% predict 4 waypoints, t+0.2, t+0.4, t+0.6, t+0.8
% predict 8 frames, t+0.1 -- t+0.9
% due to the limited gpu memory, in the third training step, we only train with the three steps. for example, input camera, action and output next action and next video as a step.
% vicuna-7b as pretrained llm, sd-1.5 as pretrained diffusion unet, animatediff as pretrained temporal module

\subsection{SoTA Comparison}

The experimental results in Table~\ref{table:sota_comparasion} demonstrate that our method significantly outperforms existing state-of-the-art approaches on the LangAuto benchmark. 
During testing, we append the most recent sensor history (up to $T_{\text{max}}$ frames) at every timestep to curb cumulative error and maintain temporal coherence. 
Specifically, LMDrive~\citep{shao2024lmdrive} achieves a driving score (DS) of 10.7 in the LangAuto track, while AD-H~\citep{zhang2024ad} and BEVDriver~\citep{winter2025bevdriver} demonstrate improved DS values of 44.0 and 48.9, respectively. Our method further pushes the performance to a higher level, with a DS of 62.2. In terms of route completion (RC) and infraction score (IS), our approach also achieves superior performance across all three tracks: LangAuto, LangAuto-Short, and LangAuto-Tiny, showing the effectiveness of our method in handling more complex driving scenarios with language instructions.

\subsection{Ablation studies}

\begin{figure*}[]
    \centering
    \begin{minipage}[t]{1.0\textwidth}
    \includegraphics[width=\linewidth]{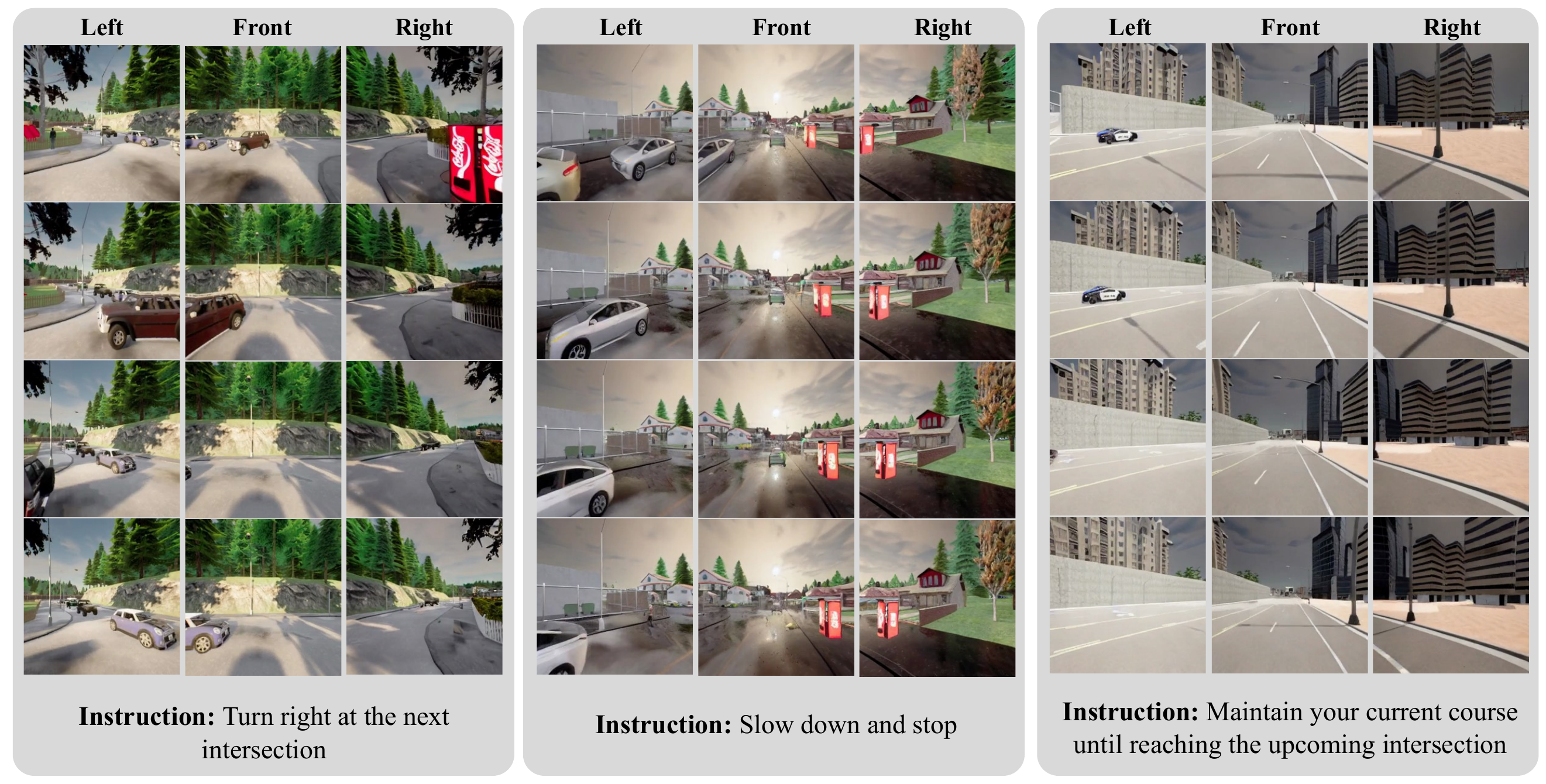}
    \end{minipage}

    \caption{Visualization of multi-view future scenes generated by LMGenDrive, showing consistent left-front-right views aligned with driving instructions.}
    \label{fig:visualization}
\end{figure*}

\noindent\textbf{Ablation Study on Module Design.} As shown in Table~\ref{tab:ablation-planning}, we conduct four ablation experiments to quantify the contribution of each key component in our proposed LMGenDrive.
(1) \textbf{w/o world generator}: Removing the world generator together with its world query sharply degrades DS to 53.4, demonstrating that the multi-view world generator is crucial for enriching the LLM’s understanding of spatio-temporal dynamics and strengthening future-scene reasoning. With a stronger understanding of world semantics, the driving agent exhibits markedly improved instruction-following capability and safer decision-making behavior.
(2) \textbf{w/o action queries}: Replacing learnable action queries with an LMDrive-style autoregressive action prediction lowers DS to 58.7, indicating that explicit action queries provide more structured supervision and lead to more reliable planning.  
(3) \textbf{w/o visual pre-training}: Without the first-stage driving-oriented visual pre-training, DS drops to 54.9, highlighting the importance of injecting driving-specific semantics into the vision encoder to enhance downstream scene understanding.  
(4) \textbf{w/o stage-3 training}: Skipping the Multi-Step Long-Horizon training stage reduces DS to 55.6, confirming that long-horizon temporal modeling is essential for robust reasoning over extended driving contexts.  
Overall, all ablations show degraded DS, with missing world generator or long-horizon training causing the largest drops, confirming these modules—along with visual pre-training and action queries—are vital for accurate, safe planning in LMGenDrive.

\noindent\textbf{Ablation Study on Generation Module Design.}
As shown in Table~\ref{tab:ablation-generation}, we further investigate how key components of the generation module influence video quality, measured by FID and FVD.  
(1) \textbf{w/o multi-view fusion}: Removing the cross-view fusion increases FID from 6.3 to 7.8 and FVD from 286 to 371, indicating that, without interaction among different camera views, the model struggles to maintain spatial consistency.  
(2) \textbf{world queries choices}: Reducing the number of world queries from 64 to 32 or 16 leads to a clear performance drop, with FID/FVD rising to 10.1/318 and 11.6/424, respectively.  
Overall, these results demonstrate that multi-view fusion and sufficient world queries are critical for generating coherent, high-quality videos.

\noindent\textbf{Ablation Study on training stages.}
To further analyze the contribution of the multi-stage training strategy, we conduct
a controlled ablation where the total number of optimization iterations is fixed to
30k across different training configurations. Specifically, we compare three settings:
(1) the full training pipeline, (2) removing Stage-2 and training Stage-3 for 30k
iterations, and (3) removing Stage-3 and training Stage-2 for 30k iterations.
As shown in Table~\ref{tab:stage_ablation_controlled}, removing Stage-3 results in a
larger performance drop on LangAuto-Short compared to LangAuto-Tiny, suggesting that
multi-step training is particularly beneficial for longer driving routes. In contrast,
Stage-2 mainly improves the stability of short-horizon grounding and action prediction.

\begin{table*}[t]
\centering
\caption{Training-stage ablation under controlled iterations.
We report DS/RC/IS on LangAuto-Tiny and LangAuto-Short.}
\label{tab:stage_ablation_controlled}
\begin{tabular}{l|ccc|ccc}
\hline
\multirow{2}{*}{\textbf{Setting}} 
& \multicolumn{3}{c|}{LangAuto-Tiny} 
& \multicolumn{3}{c}{LangAuto-Short} \\
 & DS $\uparrow$ & RC $\uparrow$ & IS $\uparrow$
 & DS $\uparrow$ & RC $\uparrow$ & IS $\uparrow$ \\
\hline
baseline (Stage-2: 2w iter, Stage-3: 1w iter) 
& \textbf{84.1} & \textbf{92.5} & \textbf{0.92} 
& \textbf{77.1} & \textbf{87.9} & \textbf{0.88} \\
w/o Stage-2 (Stage-3: 3w iter) 
& 78.0 & 87.6 & 0.90 
& 67.9 & 81.0 & 0.85 \\
w/o Stage-3 (Stage-2: 3w iter) 
& 80.1 & 88.5 & 0.91 
& 72.3 & 83.7 & 0.86 \\
\hline
\end{tabular}
\end{table*}

\subsection{Long-Horizon Video Generation Scaling}
\label{sec:long_horizon_scaling}

To evaluate the scalability of our diffusion-based world model to longer horizons,
we assess generation quality under increasing prediction lengths. Specifically,
we autoregressively generate multi-view future videos with horizons of
$\{8, 16, 24, 32, 64, 128\}$ frames and measure Fr\'echet Inception Distance (FID) and
Fr\'echet Video Distance (FVD) against ground-truth videos on the same
evaluation split, where lower values indicate better quality.

Table~\ref{tab:long_horizon_scaling} shows a clear performance degradation as the
prediction horizon increases, which is expected in autoregressive generation
where small errors accumulate over time. In long-horizon videos, we observe
three common issues: temporal drift of distant objects, motion inconsistency in
agent trajectories, and gradual deviation from the intended instruction.
Notably, FVD increases faster than FID, indicating that temporal coherence is
more sensitive to horizon length than per-frame visual quality. Overall,
LMGenDrive maintains strong fidelity for short horizons (8--32 frames), while
longer videos (64--128 frames) remain challenging due to error accumulation.

\begin{table}[t]
\centering
\setlength{\tabcolsep}{6pt}
\caption{Scaling of long-horizon video generation. FID/FVD degrade with longer prediction horizons due to accumulated errors.}
\label{tab:long_horizon_scaling}

\resizebox{0.45\textwidth}{!}{
\begin{tabular}{c|cccccc}
\hline
\textbf{Horizon (frames)} & 8 & 16 & 24 & 32 & 64 & 128 \\
\hline
FID $\downarrow$ & 6.3 & 7.4 & 7.8 & 9.2 & 12.8 & 15.1\\
FVD $\downarrow$ & 286 & 340 & 371 & 435 & 610 & 981\\
\hline
\end{tabular}
}
\end{table}

\subsection{Visualization}
To illustrate LMGenDrive’s capabilities, Figure~\ref{fig:visualization} presents qualitative videos from the CARLA simulator. The top row shows the initial multi-view observations as the conditioning inputs, while the subsequent rows visualize three future steps generated by our multi-view world model in an autoregressive manner. At each step, the model takes the previously generated multi-view frames and predicts actions as input, to synthesize the next set of left, front, and right camera views. Each panel displays these synchronized camera views together with the corresponding driving instruction. The results show that LMGenDrive (1) preserves spatial consistency across views, (2) anticipates dynamic agents such as crossing vehicles and pedestrians, and (3) aligns future scene evolution with the given language instructions. 
% These examples demonstrate how realistic, step-by-step video generation strengthens the LLM’s reasoning and supports robust long-horizon closed-loop driving.

\section{Conclusion}
We introduced LMGenDrive, a unified framework that combines LLM-based multimodal reasoning with generative world models for closed-loop end-to-end driving. By jointly modeling instruction following, spatio-temporal understanding, and multi-view future video generation, the system achieves significantly improved robustness and performance on the CARLA LangAuto benchmark. Ablation studies confirm that each component—visual pretraining, action and world queries, long-horizon training, and multi-view generation—plays an essential role.
Our results demonstrate the complementary strengths of integrating understanding and generation within a single architecture, enabling more anticipatory and reliable behavior than prior approaches. Looking forward, LMGenDrive provides a foundation for scaling unified reasoning–generation models to real-world settings and for exploring broader applications in embodied decision-making.

\bibliographystyle{splncs04}
\bibliography{main}
\end{document}